\def\year{2019}\relax
\documentclass[letterpaper]{article} 
\usepackage{aaai19}  
\usepackage{times}  
\usepackage{helvet}  
\usepackage{courier}  
\usepackage{url}  
\usepackage{graphicx}  
\usepackage{latexsym}
\usepackage{amsfonts}
\usepackage{braket}
\usepackage{bm}
\usepackage{amsmath}
\usepackage{multirow}
\usepackage{amsthm}
\begin{document}
%
\title{A Generalized Language Model in Tensor Space}
\author{Lipeng Zhang$^\dagger$, Peng Zhang$^\dagger$\thanks{Corresponding author: Peng Zhang (pzhang@tju.edu.cn)}, Xindian Ma$^\dagger$, Shuqin Gu$^\dagger$, Zhan Su$^\dagger$, Dawei Song$^\ddagger$\\
$^\dagger$College of Intelligence and Computing, Tianjin University, Tianjin, China\\
$^\ddagger$School of Computer Science and Technology, Beijing Institute of Technology, Beijing, China\\
\{lpzhang, pzhang, xindianma, shuqingu, suzhan\}@tju.edu.cn, dawei.song2010@gmail.com\\
}
\maketitle
\begin{abstract}
In the literature, tensors have been effectively used for capturing the context information in language models. However, the existing methods usually adopt relatively-low order tensors, which have limited expressive power in modeling language. Developing a higher-order tensor representation is challenging, in terms of deriving an effective solution and showing its generality. In this paper, we propose a language model named Tensor Space Language Model (TSLM), by utilizing tensor networks and tensor decomposition. In TSLM, we build a high-dimensional semantic space constructed by the tensor product of word vectors. Theoretically, we prove that such tensor representation is a generalization of the $n$-gram language model. We further show that this high-order tensor representation can be decomposed to a recursive calculation of conditional probability for language modeling. The experimental results on Penn Tree Bank (PTB) dataset and WikiText benchmark demonstrate the effectiveness of TSLM.
\end{abstract}

\section{Introduction}
Language Modeling (LM) is a fundamental research topic that underpins a wide range of Natural Language Processing (NLP) tasks, e.g., speech recognition, machine translation, and dialog system~\cite{Yu2014Automatic,Lopez2008Statistical,Wang2006Automatic}. Statistical learning of a language model aims to approximate the probability distribution on the set of expressions in the language~\cite{Brown1992Class}. Recently, Neural networks (NNs), e.g., Recurrent Neural Networks (RNNs) have been shown effective for modeling language~\cite{BengioDVJ03,Mikolov2010Recurrent,jozefowicz2016exploring}. 

\begin{figure*}[htp]
\centering
\includegraphics[scale=0.40]{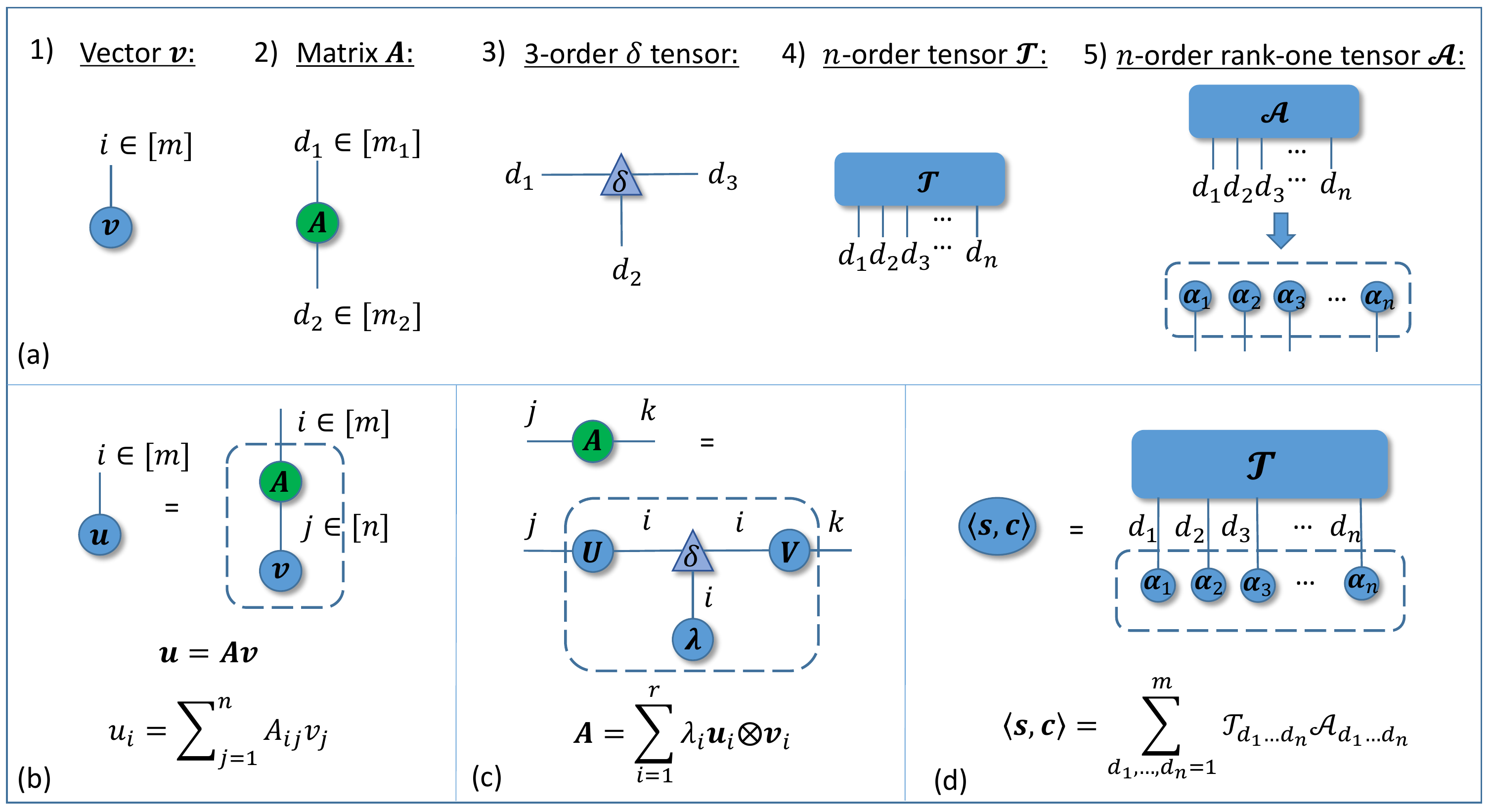}
\caption{Introduction of Tensor Networks. (a) Tensors in the TN are represented by nodes. (b) A matrix $A$ multiplying a vector $\bm{v}$ in TN notation. The contracted indices are denoted by $j$ and are summed upon. The open indices are denoted by $i$, their number equals the order of the tensor represented by the entire network. The contraction is marked by the dashed line. (c) A TN notation illustrates the SVD of a matrix $A$. (d) A TN notation shows the inner product of two tensors.}
\label{fig:figure1}
\end{figure*}


In the literature, a dense tensor is often used to represent a sentence or document. Cai et al.~\shortcite{cai2006tensor} proposed TensorLSI, which considered sentences or documents as $2$-order tensors (matrices) and tried to find an optimal basis for the tensor subspace in term of reconstruction error. The $2$-order tensor (matrix) only reflects the local information (e.g., bi-gram information). Liu et al.~\shortcite{Liu2005Text} proposed to model text by a multilinear algebraic tensor instead of a vector. Specifically, they represented texts using $3$-order tensors to capture the context of words. However, they still adopted relatively low-order tensors, rather than high-order ones constructed by the tensor product of vectors. For a sentence with $n$ words ($n>3$), a $n$-order tensor (a high-order tensor) constructed by the tensor product of $n$ word vectors, can consider all the combinatorial dependencies among words (not limited to two/three consecutive words in 2/3-order tensor). It turns out that low-order
tensors have limited expressive power in modeling language.

It is challenging to construct a high-order tensor based language model (LM), in the sense that it will involve an exponentially increasing number of parameters. Therefore, the research problems are how to derive an effective high-order tensor based LM and how to demonstrate the generality of such a tensor-based LM. To address these problems, our motivation is to explore the expressive ability of tensor space in depth, by making use of tensor networks and tensor decomposition, in the language modeling process. 

Tensor network is an elegant mathematical tool and an effective method for solving high-order tensors (e.g., tensors in quantum many-body problem), through contractions among lower-order tensors~\cite{Pellionisz1980Tensorial}. Recent research shows the connections between neural networks and tensor networks, which provides a novel perspective for the interpretation of neural network (Cohen et al.~\shortcite{Cohen2016On} and Levine et al.~\shortcite{levine2017benefits,levine2018deep}). With the help of tensor decomposition, the high dimensionality of parameters in tensor space can be reduced greatly. Based on tensors, a novel language representation using quantum many-body wave function  was also proposed in~\cite{zhang2018quantum}.



Inspired by the recent research, in this paper, we propose a Tensor Space Language Model (TSLM), which is based on high-dimensional semantic space constructed by the tensor product of word vectors. Tensor operations (e.g., multiplication, inner product, decomposition) can be represented intuitively in tensor networks. Then, the probability of a sentence will be obtained by the inner product of two tensors, corresponding to the input data and the global parameters, respectively. 

Theoretically, we prove that TSLM is a generalization of the $n$-gram language model, by showing that the conditional probability of a language sequence can be calculated by the tensor representation we constructed. We further show that, after tensor decomposition, the high-order tensor representation can be decomposed to a recursive calculation of conditional probability for language modeling. Finally, we evaluate our model on the PTB dataset and the WikiText benchmark, and the experimental results demonstrate the effectiveness of TSLM.

The main contributions of our work can be summarized as follows: (1) We propose a novel language model aiming to consider high-order dependencies of words via tensors and tensor networks. (2) We prove that TSLM is a generalization of the $n$-gram language model. (3) We can derive a recursive calculation of conditional probability for language modeling via tensor decomposition in TSLM.

\section{Preliminaries}
\label{sect:2}

We first briefly introduce tensors and tensor networks.
\subsection{Tensors}
$\bm{1.}$ A {\bf tensor} can be thought as a multidimensional array. The {\bf order} of a tensor is defined to be the number of indexing entries in the array, which are referred to as modes. The {\bf dimension} of a tensor in a particular mode is defined as the number of values that may be taken by the index in that mode. A tensor $\mathcal{T} \in \mathbb{R}^{m_1\times\cdots\times m_n}$ means that it is a $n$-order tensor with dimension $m_i$ in each mode $i\in[n]:=\{1,\ldots,n\}$. For simplicity, we also call it a $m^n$-dimensional tensor in the following text. A specific entry in a tensor will be referenced with subscripts, e.g. $\mathcal{T}_{d_1\ldots d_n}\in\mathbb{R}.$

\noindent$\bm{2.}$ {\bf Tensor product} is a fundamental operator in tensor analysis, denoted by $\otimes$, which can map two low-order tensors to a high-order tensor. Similarly, the tensor product of two vector spaces is a high-dimensional tensor space. For example, tensor product intakes two tensors $\mathcal{A}\in\mathbb{R}^{m_1\times\cdots\times m_j}$ ($j$-order) and $\mathcal{B}\in\mathbb{R}^{m_{j+1}\times\cdots\times m_{j+k}}$ ($k$-order), and returns a tensor $\mathcal{A}\otimes\mathcal{B}=\mathcal{T}\in\mathbb{R}^{m_1\times\cdots\times m_{j+k}}$ (($j$+$k$)-order) defined by : $\mathcal{T}_{d_1\ldots d_{j+k}}=\mathcal{A}_{d_1\ldots d_j}\cdot\mathcal{B}_{d_{j+1}\ldots d_{j+k}}$.

\noindent$\bm{3.}$ A $n$-order tensor $\mathcal{A}$ is {\bf rank-one} if it can be written as the tensor product of $n$ vectors, i.e.,
\begin{equation}
\label{rank-1}
\mathcal{A}=\bm{\alpha}_1\otimes\bm{\alpha}_2\otimes\cdots\otimes\bm{\alpha}_n
\end{equation}
This means that each entry of the tensor is the product of the corresponding vector elements:
\begin{equation}
\label{rank-1-element}
\mathcal{A}_{d_1d_2\ldots d_n}=\alpha_{1,d_1}\alpha_{2,d_2}\cdots \alpha_{n,d_n}~~~\forall~i, d_i\in[m_i]
\end{equation}

\noindent$\bm{4.}$ The {\bf rank} of a tensor $\mathcal{T}$ is defined as the smallest number of rank-one tensors that generate $\mathcal{T}$ as their sum~\cite{Hitchcock1927The,Kolda2009Tensor}.

\subsection{Tensor Networks}
A {\bf Tensor Network} (TN) is formally represented by an undirected and weighted graph. The basic building blocks of a TN are tensors, which are represented by nodes in the network. The order of a tensor is equal to the number of edges incident to it. The weight of a edge is equal to the dimension of the corresponding mode of a tensor. Fig.~\ref{fig:figure1} (a) shows five examples for tensors: 1) A vector is a node with one edge. 2) A matrix is a node with two edges. 3) In particular, a triangle node represents a $\delta~$tensor, $\delta\in\mathbb{R}^{m\times\cdots\times m}$, which is defined as follow:
\begin{equation}
\label{eq:delta}
\delta_{d_1\ldots d_n}=\left\{
             \begin{array}{lr}
             1,~~~& d_1=\cdots=d_n   \\
             0,~~~& \mathrm{otherwise}  
             \end{array}
\right.
\end{equation}
with $d_i\in[m]~\forall i\in[n]$, i.e. its entries are equal to one only on the super-diagonal and zero otherwise. 4) The rounded rectangle node is the same as the circle node representing an arbitrary tensor. Accordingly, a tensor of order $n$ is represented in the TN as a node with $n$ edges. 5) A $n$-order rank-one tensor $\mathcal{A}$ can be represented by the tensor product of $n$ vectors.

Edges which connect two nodes in the TN represent a contraction operation between the two corresponding modes. An example for a contraction is depicted in Fig.~\ref{fig:figure1} (b), in which a TN corresponding to the operation of multiplying a vector $\bm{v}\in\mathbb{R}^{n}$ by a matrix $A\in\mathbb{R}^{m\times n}$ is performed by summing over the only contracted index $j$. As there is only one open index $i$, the result of contracting the network is a vector: $\bm{u}\in\mathbb{R}^{m}$ which upholds $\bm{u}=A\bm{v}$. 

An important concept in the later analysis is Singular Value Decomposition (SVD), which denotes that a matrix $A\in\mathbb{R}^{m\times n}$ can be decomposed as $A=U\Lambda V$, where $\Lambda\in\mathbb{R}^{r\times r}$ represents a diagonal matrix and $U\in\mathbb{R}^{m\times r}, V\in\mathbb{R}^{r\times n}$ represent orthogonal matrices. In TN notation, we represent the SVD as $A=\sum_{i=1}^{r}\lambda_{i}\bm{u}_{i}\otimes\bm{v}_{i}$ in Fig.~\ref{fig:figure1} (c). $\lambda_i$ is a singular value, $\bm{u}_{i}$, $\bm{v}_{i}$ are the components of $U$, $V$ respectively. The effect of the $\delta$ tensor is shown obviously, which can be observed as `forcing' the $i$-th `row' of any tensor to be multiplied only by the $i$-th `rows' of other tensors.

Fig.~\ref{fig:figure1} (d) shows the {\bf inner product} of two tensors $\mathcal{T}$ and $\mathcal{A}$, which returns a scalar value that is the sum of the products of their entries~\cite{Kolda2009Tensor}.

\section{Basic Formulation of Language Modeling in Tensor Space}
\label{sect:3}
In this section, we describe the basic formation of Tensor Space Language Model (TSLM), which is used for computing probability for the occurrence of a sequence $s=w_1^n:=(w_1,\ldots,w_n)$ of length $n$, composed of $|\mathrm{V}|$ different tokens, and the vocabulary $\mathrm{V}$ containing all the words in the model, i.e. $w_i\in \mathrm{V}$.  In the next sections, we will prove that when we use one-hot vectors, TSLM is a generalization of $n$-gram language model. If using word embedding vectors, TSLM can result in a recursive calculation of conditional probability for language modeling. In Related Work, we will discuss the relations and differences between our previous work on language representation for matching sentences \cite{zhang2018quantum} and the tensor space language modeling in this paper. 



\subsection{Representations of Words and Sentences}

First, we define the semantic space of a single word as a $m$-dimensional vector space $\mathbb{V}$ with the orthogonal basis $\{\bm{e}_d\}_{d=1}^m$, where each base vector $\bm{e}_{d}$ is corresponding to a specific semantic meaning. A word $w_i$ in a sentence $s$ can be written as a linear combination of the $m$ orthogonal basis vectors as a general representation:
\begin{equation}
\label{eq:singlewave}
\bm{w}_i=\sum_{d_i=1}^{m}\alpha_{i,d_i}\bm{e}_{d_i}
\end{equation}
where $\alpha_{i,d_i}$ is its corresponding coefficient. For the basis vectors $\{\bm{e}_{d}\}_{{d}=1}^m$, there can be two different choices, one-hot vectors or embedded vectors.

As for a sentence $s=(w_1,\ldots,w_n)$ with length $n$, it can be represented in the tensor space:
\begin{equation}
\label{eq:Hilbert_n}
\mathbb{V}^{\otimes n}:= \underbrace{\mathbb{V}\otimes\mathbb{V}\otimes\cdots\otimes\mathbb{V}}_n
\end{equation}
as:
\begin{equation}
\label{eq:Seq_Words}
\bm{s}=\bm{w}_{1}\otimes\cdots\otimes\bm{w}_{n}
\end{equation}
which is a $m^n$-dimensional tensor. Through the interaction of each dimension of words, the sentence tensor has a strong expressive power.  Substituting Eq.~\ref{eq:singlewave} into Eq.~\ref{eq:Seq_Words}, the representation of the sentence $\bm{s}$ can be expanded by:
\begin{equation}
\label{eq:text-rep-ps}
\bm{s}=\sum_{d_1,\ldots,d_n=1}^{m}\mathcal{A}_{d_1\ldots d_n}\bm{e}_{d_1}\otimes\cdots\otimes\bm{e}_{d_n}
\end{equation}
where $\{\bm{e}_{d_1}\otimes\cdots\otimes \bm{e}_{d_n}\}_{d_1,\ldots,d_n=1}^{m}$ are the basis with $m^n$ dimension in the tensor space $\mathbb{V}^{\otimes n}$, which denotes the high-dimensional semantic meaning. $\mathcal{A}$ is a $m^n$-dimensional tensor and its each entry ${\mathcal{A}_{d_1\ldots d_n}}$ is the corresponding coefficient of each basis. According to the Eq.~\ref{rank-1} and \ref{rank-1-element}, we will see $\mathcal{A}$, computed as $\prod_{i=1}^{n}\alpha_{i,d_i}$, is a rank-one tensor.

\subsection{A Probability Estimation Method of Sentences}
A goal of language modeling is to learn the joint probability function of sequences of words in a language~\cite{BengioDVJ03}. We assume that each sentence $s_i$ appears with a probability $p_i$. Then, we can construct a mixed representation $c$ which is a linear combination of sentences, denoted as:
\begin{equation}
\label{eq:text_rep_c}
\bm{c}:= \sum p_i\bm{s}_i
\end{equation}

We consider the mixed representation $\bm{c}$ as a high-dimensional representation of a sequence containing $n$ words. For each sentence $s_i$, it can be represented by coefficient tensor $\mathcal{A}_i$ and basis vectors. Therefore, based on Eqs.~\ref{eq:text-rep-ps} and~\ref{eq:text_rep_c}, the mixed representation $\bm{c}$ can be formulated  with a coefficient tensor $\mathcal{T}$:
\begin{equation}
\label{eq:text-rep}
\bm{c} = \sum_{d_1,\ldots,d_n=1}^{m}\mathcal{T}_{d_1\ldots d_n} \bm{e}_{d_1}\otimes\ldots\otimes\bm{e}_{d_n}\\
\end{equation}

The difference between $\bm{s}$ in Eq.~\ref{eq:text-rep-ps} and $\bm{c}$ in Eq.~\ref{eq:text-rep} is mainly on two different tensors $\mathcal{A}$ and $\mathcal{T}$. According to the {\bf Preliminaries},  $\mathcal{A}$ is essentially rank-one while $\mathcal{T}$ has a higher rank and is harder to solve. We will show that the tensor $\mathcal{T}$ encodes the parameters, and $\mathcal{A}$ is the input, in our model.

In turn, if we have estimated such a mixed representation $\bm{c}$ (in fact, its coefficient tensor $\mathcal{T}$) from our model, we can get the probability $p_i$ of a sentence $s_i$ via computing the inner product of $\bm{c}$ and $\bm{s}_i$:
\begin{equation}
\label{eq:text_rep_in}
p(s_i)=\langle\bm{s}_i,\bm{c}\rangle
\end{equation}

Based on the Eq.~\ref{eq:text-rep-ps},~\ref{eq:text-rep} and~\ref{eq:text_rep_in}, we can obtain:
\begin{equation}
\label{eq:projection}
\begin{aligned}
p(s_i)&= \sum_{d_1,\ldots,d_n=1}^{m}\mathcal{T}_{d_1\ldots d_n}\mathcal{A}_{d_1\ldots d_n}\\
\end{aligned}
\end{equation}
as shown in the tensor network of Fig~\ref{fig:figure1}(d). This is the basic formula in 
TSLM for estimating the sentence probability.

\section{TSLM as a Generalization of N-Gram Language Model}
\begin{figure*}[t]
\centering
\includegraphics[scale=0.48]{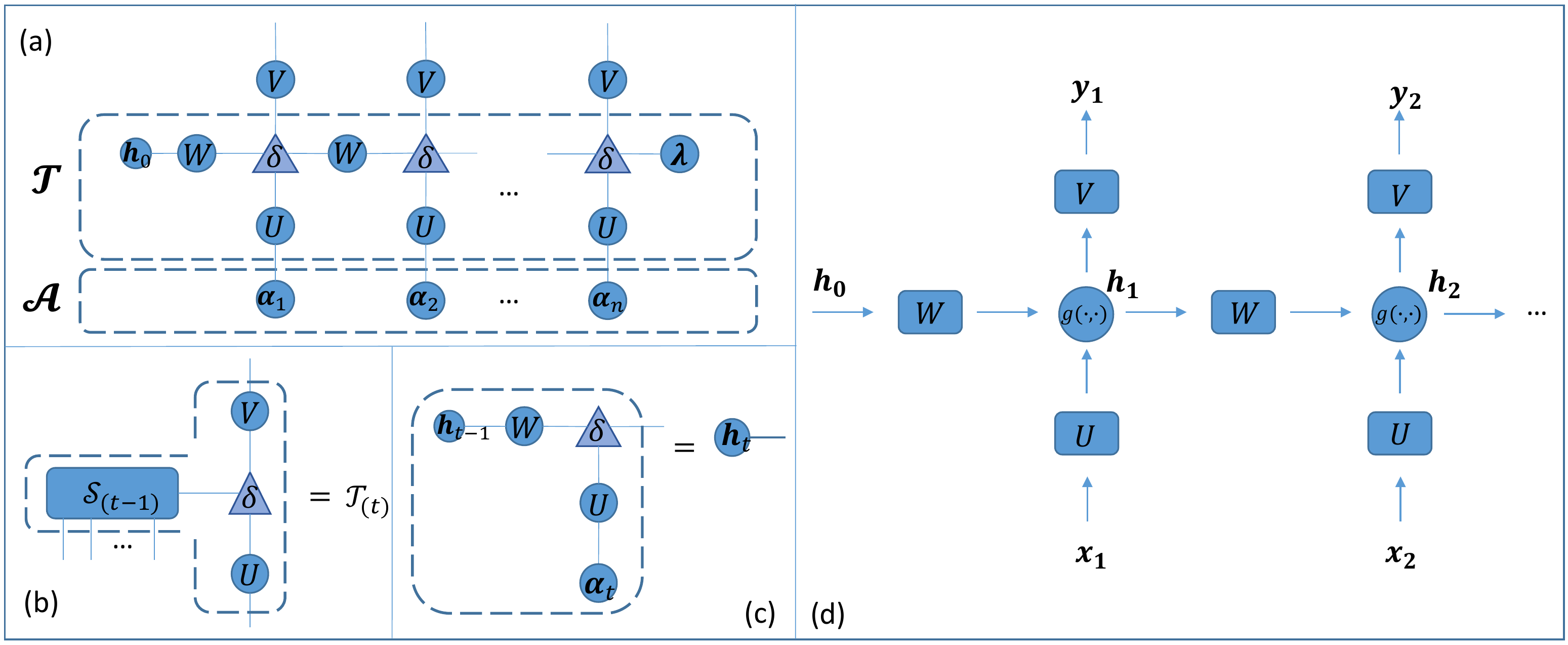}
\caption{The TN represents the recursive calculation process of TSLM. (a) represents the inner product of two tensors $\mathcal{T}$ and $\mathcal{A}$, (b) denotes the recursive representations of $\mathcal{T}_{(t)}$ and (c) is the recursive representations of $\bm{h}_t$, respectively. (d) is a general RNN architecture.}
\label{fig:figure3}
\end{figure*}
The goal of the $n$-gram language model is to estimate the probability distribution of sentences. For a specific sentence $s$, its joint probability $p(s)=p(w_{1}^{n}):=p(w_{1},\ldots,w_n)$ relies on the Markov Chain Rule of conditional probability: 
\begin{equation}
p(w_{1}^{n})=p(w_1)\prod_{i=2}^{n}p(w_i|w_{1}^{i-1})
\end{equation}
and the conditional probability $p(w_i|w_1^{i-1})$ can be calculated as:
\begin{equation}
\begin{aligned}
p(w_i|w_{1}^{i-1})&=\frac{p(w_{1}^{i})}{p(w_{1}^{i-1})}\approx \frac{count(w_{1}^{i})}{count(w_{1}^{i-1})}\\
\end{aligned}
\end{equation}
where the $count$ denotes the frequency statistics in corpus. 

\noindent{\bf Claim 1.} In our TSLM, when we set the dimension of vector space $m=|\mathrm{V}|$ and each word $w$ as an {\it one-hot} vector, the probability of sentence $s$ consist of words $d_1\ldots d_n$ in vocabulary is the entry $\mathcal{T}_{d_1\ldots d_n}$ of tensor $\mathcal{T}$.
\begin{proof}
The detailed proof can be found in Appendix.
\end{proof}
Intuitively, the specific sentence $s$ can be represented as an one-hot tensor. The mixed representation $\bm{c}$ can be regarded as the total sampling distribution. The tensor inner product $\langle \bm{s},\bm{c}\rangle$ represents statistics probability that a sentence $s$ appears in a language.

\noindent{\bf Claim 2.}
\label{claim2}
In our TSLM, we define the word sequence $w_1^i=(w_1,w_2,\ldots,w_i)$ with length $i$ as:
\begin{equation}
\label{eq:w1i}
\bm{w}_{1}^{i} := \bm{w}_{1}\otimes\cdots\otimes\bm{w}_{i}\otimes\bm{1}_{i+1}\otimes\cdots\otimes\bm{1}_n
\end{equation}
which means that the sequence $w_{1}^{i}$ is padded via using full one vector $\bm{1}$. Then, the probability $p(w_1^i)$ can be computed as $p(w_1^i)=\langle \bm{w}_1^i,\bm{c}\rangle$.
\begin{proof}
It can be proved by the marginal probability of multiple discrete variables (in Appendix).
\end{proof}
Therefore, the conditional probability $p(w_i|w_1^{i-1})$ in $n$-gram language model can be calculated by Bayesian Conditional Probability Formula using tensor representations and tensor inner product as follow:
\begin{equation}
\label{eq:n-gram}
p(w_i|w_1^{i-1}) = \frac{\langle \bm{w}_1^i,\bm{c}\rangle}{\langle\bm{w}_1^{i-1},\bm{c}\rangle}
\end{equation}
This kind of representation of conditional probability is a special case of TSLM based on the one-hot basis vectors. Because of the one-hot vector representation, the $n$-gram language model has $O(|\mathrm{V}|^{n})$ parameters. Compared with $n$-gram language model, 
our general model has $O(m^{n})$ parameters ($m\ll |\mathrm{V}|$). However, the tensor space still contains exponential parameters, and in the next section, we will introduce tensor decomposition to deal with this problem.

\section{Deriving Recursive Language Modeling Process from TSLM}
\label{sect:4}
We have defined the method for estimating probability of a sentence $s$ by the inner product of two tensors $\mathcal{T}$ and $\mathcal{A}$ in Eq.~\ref{eq:projection}. In this section, we describe the recursive language modeling process in Fig.~\ref{fig:figure3}. Firstly, our derivation is under the condition of basis vectors $\{\bm{e}_{d}\}_{{d}=1}^m$ as embedded vectors. 
Secondly, we recursively decompose the tensor $\mathcal{T}$ (see Fig.~\ref{fig:figure2}), to obtain the TN representation of tensor $\mathcal{T}$ in Fig.~\ref{fig:figure3} (a). Then, we use the intermediate variables in Fig.~\ref{fig:figure3} (bc) to estimate the conditional probability $p(w_t|w_1^{t-1})$. In the following, we introduce the recursive tensor decomposition, followed by the calculation of conditional probability.

\subsection{Recursive Tensor Decomposition}
\begin{figure}[t]
\centering
\includegraphics[scale=0.32]{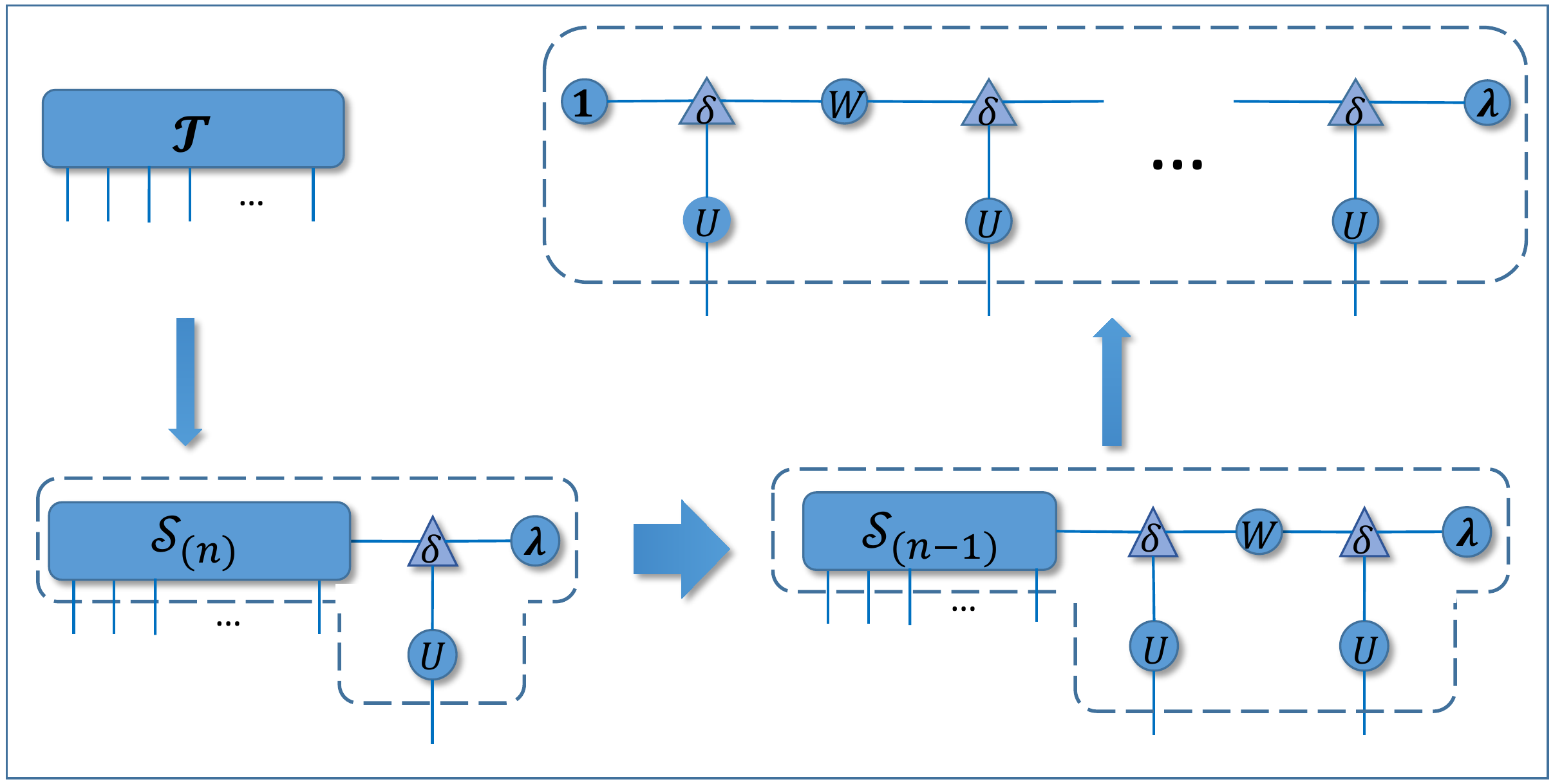}
\caption{ The recursive generalized SVD for the tensor $\mathcal{T}$.}
\label{fig:figure2}
\end{figure}

We generalize the SVD from matrix to tensor as shown intuitively in Fig.~\ref{fig:figure2}, inspired by the train-style decomposition of Tensor-Train~\cite{Oseledets2011Tensor} and Tucker Decomposition~\cite{Kolda2009Tensor}. 
Fig.~\ref{fig:figure2} illustrates a high-order tensor recursively decomposed as several low-order tensor (vectors, matrices, etc.).

The formula of the recursive decomposition about tensor $\mathcal{T}$ is :
\begin{equation}
\begin{aligned}
\label{eq:general_SVD}
\mathcal{T}&=\sum_{i=1}^{r}\lambda_i\mathcal{S}_{(n),i}\otimes\bm{u}_i \\
\mathcal{S}_{(n),k}&=\sum_{i=1}^{r}W_{k,i}\mathcal{S}_{(n-1),i}\otimes\bm{u}_i \\
\end{aligned}
\end{equation}
where we define : $\mathcal{S}_{(1)}=\bm{1}\in\mathbb{R}^{r}$. It means that a $n$-order tensor $\mathcal{T}\in\mathbb{R}^{m\times\cdots\times m}$ can be decomposed as a $n$-order tensor\footnote{To distinguish, we use the first parenthesized subscript to indicate the order of the tensor.} $\mathcal{S}_{(n)}\in\mathbb{R}^{m\times\cdots\times r}$, a diagonal matrix $\Lambda\in\mathbb{R}^{r\times r}$ and a matrix $U\in\mathbb{R}^{r\times m}$. One can consider this decomposition as the matrix SVD after tensor matricization, also as unfolding or flattening the tensor to matrix by one mode. Recursively, ($n$-$1$)-order tensor $\mathcal{S}_{(n),k}$, which can be seen as the $k$-th `row' of the tensor $\mathcal{S}_{(n)}$, can be decomposed like the tensor $\mathcal{T}$ and $W$ is a matrix composed by $r$ groups of singular value vectors.

We employ this decomposition to extract the main features of the tensor $\mathcal{T}$ which is similar with the effect of SVD on matrix, then approximately represent the parameters of our model, where $r$ ($r\leq m$) denotes the rank of tensor decomposition. This tensor decomposition method reduce the $O(m^n)$ magnitude of parameters approximatively to $O(m\times r)$.

\subsection{A Recursive Calculation of Conditional Probability}
\label{sect:RNNLM}
In our model, we compute the conditional probability distribution as :
\begin{equation}
\label{conditional_probability}
p(w_t|w_1^{t-1}) = softmax(\langle\mathcal{T}_{(t)},\mathcal{A}_{(t-1)}\rangle)
\end{equation}
where $\mathcal{A}_{(t-1)}$ is the input of $(t$-$1)$ words, represented as $\bm{\alpha}_1,\ldots,\bm{\alpha}_{(t-1)}$ in Fig.~\ref{fig:figure3} (a). $\langle\mathcal{T}_{(t)},\mathcal{A}_{(t-1)}\rangle$ is denoted as $\bm{y}_t$.

As shown in Fig.~\ref{fig:figure3} (b), $\mathcal{T}_{(t)}\in\mathbb{R}^{m\times\cdots\times m\times|\mathrm{V}|}$ is constructed by matrix $V\in\mathbb{R}^{r\times|\mathrm{V}|}$, $\mathcal{S}_{(t-1)}$ and matrix $U$. The $V$ is the weighted matrix mapping to the vocabulary: 
\begin{equation}
\label{Tt}
\mathcal{T}_{(t),k} = \sum_{i=1}^{r}V_{k,i}\mathcal{S}_{(t-1),i}\otimes\bm{u}_i
\end{equation}

As shown in Fig.~\ref{fig:figure3} (c), when calculating the inner product of two tensors $\mathcal{T}$ and $\mathcal{A}$, we can introduce intermediate variables $\bm{h}_t$, which can be recursively calculated as:
\begin{equation}
\label{ht}
\begin{aligned}
\bm{h}_1 &= W\bm{h}_{0}\odot U\bm{\alpha}_1\\
...\\
\bm{h}_t &= W\bm{h}_{t-1}\odot U\bm{\alpha}_t\\
\end{aligned}
\end{equation}
where setting the $\bm{h}_0:= W^{-1}\bm{1}$, and the matrices $W$ and $U$ decomposed from Eq.~\ref{eq:general_SVD} in the last section. The symbol $\odot$ denotes the element-wise multiplication between vectors. This multiplicative operation is derived from the tensor product (in Eq.~\ref{eq:general_SVD}) and the $\delta$-tensor, which we have explained it `forces' the two vectors connected with it to be multiplied by elements (in Preliminaries).

Based on the analysis above, the recursive calculation of conditional probability of our TSLM can be formulated as:
\begin{equation}
\begin{aligned}
\label{eq:RNN}
p(w_t|w_1^{t-1})&=softmax(\bm{y}_t)\\
\bm{y}_t&=V\bm{h}_t\\
\bm{h}_t&=g(W\bm{h}_{t-1},U\bm{\alpha}_t)\\
g(\bm{a},\bm{b})&=\bm{a}\odot\bm{b}\\
\end{aligned}
\end{equation}
where $\bm{\alpha}_t\in\mathbb{R}^{m}$ is the input at time-step $t\in[n]$, $\bm{h}_t\in\mathbb{R}^{r}$ is the hidden state of the network, $\bm{y}_t\in\mathbb{R}^{|\mathrm{V}|}$ denotes the output, and the trainable weight parameters $U\in\mathbb{R}^{m\times r}, W\in\mathbb{R}^{r\times r}, V\in\mathbb{R}^{r\times |\mathrm{V}|}$ are the input to hidden, hidden to hidden and hidden to output weights matrices, respectively, and $g$ is a non-linear operation.

The operation $\odot$ stands for element-wise multiplication between vectors, for which the resultant vector upholds $(\bm{a}\odot\bm{b})_i=a_i\cdot b_i$. Differently, in the RNN and TSLM architecture, $g$ is defined as:
\begin{equation}
\begin{aligned}
\label{rnntslm}
g_{{ }_{RNN}}(\bm{a},\bm{b})&=\sigma(\bm{a}+\bm{b})\\
g_{{ }_{TSLM}}(\bm{a},\bm{b})&=\bm{a}\odot\bm{b}\\
\end{aligned}
\end{equation}
where $\sigma(\cdot)$ is typically a point-wise activation function such as sigmoid, tanh etc. A bias term is usually added to Eq.~\ref{rnntslm}. Since it has no effect with our analysis, we omit it for simplicity. We show a general structure of RNN in Fig.~\ref{fig:figure3}(d).

In fact, recurrent networks that include the element-wise multiplication operation have been shown to outperform many of the existing RNN models~\cite{sutskever2011generating,wu2016multiplicative}. Wu et al.~\shortcite{wu2016multiplicative} had given a more general formula for hidden unit in RNN, named Multiplicative Integration, and discussed the different structures of the hidden unit in RNN. 

\begin{table*}[t]
  
  \label{tab:dataset}
  \centering
  \begin{tabular}{ccccccc}
    \hline
    \multirow{2}{*}{}& \multicolumn{3}{c}{\bf PTB} & \multicolumn{3}{c}{\bf WikiText-2}\\
    \cline{2-7}
    &Train&Valid&Test&Train&Valid&Test\\
    \hline
    Articles &-&-&-&600&60&60\\
    Tokens   &929,590&73,761&82,431&2,088,628&217,646&245,569\\
    \hline
    Vocab size   & &10,000& & &33,278& \\
    OOV rate    & &4.8\%& & &2.6\%& \\
    \hline
  
\end{tabular}
\caption{Statistics of the PTB and WikiText-2.}
\end{table*}
\begin{table*}[t]\small
\centering
\begin{tabular}{l|cccc|cccc}
\hline
\multirow{2}{*}{}& \multicolumn{4}{c|}{\bf PTB} & \multicolumn{4}{c}{\bf WikiText-2}\\
\hline
{\bf Model} & {\bf Hidden size} & {\bf Layers} & {\bf Valid} & {\bf Test} & {\bf Hidden size} & {\bf Layers} & {\bf Valid} & {\bf Test} \\
\hline
KN-5\cite{mikolov2012context} & - & - & - & 141.2 & - & - & - & - \\
RNN\cite{mikolov2012context} & 300 & 1 & - & 124.7 & - & - & - & - \\

LSTM\cite{zaremba2014recurrent} & 200 & 2 & 120.7 & 114.5 & - & - & - & -\\
LSTM\cite{grave2016improving} & 1024 & 1 & - & 82.3 & 1024 & 1 & - & 99.3 \\
LSTM\cite{Merity2017Pointer} & 650 & 2 & 84.4 & 80.6 & 650 & 2 & 108.7 & 100.9 \\
\hline
RNN$\dagger$ & 256 & 1 & 130.3 & 124.1 & 512 & 1 & 126.0 & 120.4 \\
LSTM$\dagger$ & 256 & 1 & 118.6 & 110.3 & 512 & 1 & 105.6 & 101.4 \\
TSLM & 256 & 1 & {\bf 117.2} & {\bf 108.1} & 512 & 1 & {\bf 104.9} & {\bf 100.4} \\
RNN+MoS$\dagger$\cite{yang2018Breaking} & 256 & 1 & 88.7 & 84.3 & 512 & 1 & 85.6 & 81.8 \\  
TSLM+MoS & 256 & 1 & {\bf 86.4} & {\bf 83.6} & 512 & 1 & {\bf 83.9} & {\bf 81.0} \\
\hline
\end{tabular}
\caption{Best perplexity of models on the PTB and WikiText-2 dataset. Models tagged with $\dagger$ indicate that they are reimplemented by ourselves.}
\label{table:ptbwiki2}
\end{table*}
\section{Related Works}
\label{sect:6}
Here, we present a brief review of related work, including some representative work in language modeling, and the more recent research on the cross fields of tensor network, neural network and language modeling.

There have been tremendous research efforts in the field of statistical language modeling. Some earlier language models are based on the Markov assumption are represented by $n$-gram models \cite{Brown1992Class}, where the prediction of the next word is often conditioned just on $n$ preceding words. For $n$-gram models, Kneser and Ney~\shortcite{Kneser2002Improved} proposed the most well-known KN smoothing method, and some researchers continued to improve the smoothing method, as well as introduced the low-rank model. Neural Probabilistic Language Model~\cite{BengioDVJ03} is to learn the joint probability function of sequence of words in a language, which shows the improvement on $n$-gram models. Recently, RNN~\cite{Mikolov2010Recurrent} and Long Short-Term Memory (LSTM) networks~\cite{Soutner2013Application} achieve promising results on language model tasks. 

Recently, Cohen et al.~\shortcite{Cohen2016On} and Levine et al.~\shortcite{levine2017benefits,levine2018deep} use tensor analysis to explore the expressive power and interpretability of neural networks, including convolutional neural network (CNN) and RNN. Levine et al.~\shortcite{levine2018deep} even explored the connection between quantum entanglement and deep learning. Inspired by their work, \citeauthor{zhang2018quantum} (\citeyear{zhang2018quantum}) proposed a Quantum Many-body Wave Function inspired Language Modeling (QMWF-LM) approach. However, in QMWF-LM and TSLM, the term \textit{language modeling} has different meanings and application tasks. 
 Specifically,  QMWF-LM is basically a \textit{language representation} which encodes language features extracted by a CNN, and performs the semantic matching in Question Answer (QA) task as an \textit{extrinsic evaluation}.

Different from QMWF-LM, in this paper, TSLM focuses on the Markov process of conditional probabilities in language modeling task with an \textit{intrinsic evaluation}. Based on the tensor representation and tensor networks, we propose the tensor space language model. We have established the connection between TSLM and neural language models (e.g., RNN based LMs) and proved that TSLM is a more general language model. 

\section{Experiments}
\label{sect:5}

\subsection{Datasets}

{\bf PTB} Penn Tree Bank dataset~\cite{Marcus1993Building} is often used to evaluate language models. It consists of 929k training words, 73k validation words, 82k test words, and has 10k words in its vocabulary. 

\noindent{\bf WikiText-2 (WT2)} dataset~\cite{Merity2017Pointer}. Compared with the preprocessed version of PTB, WikiText-2 is larger. It also features a larger vocabulary and retains the original case, punctuation and numbers, all of which are removed in PTB. It is composed of full articles.

Table 1 shows statistics of these two datasets. The out of vocabulary (OOV) rate denotes the percentage of tokens have been replaced by an $\langle unk\rangle$ token. The token count includes newlines which add to  the WikiText-2 dataset.

\subsection{Evaluation Metrics}
{\bf Perplexity} is the typical measure used for reporting progress in language modeling. It is the average per-word log-probability on the holdout data set.
$$ \mathrm{PPL} = e^{(-\frac{1}{n}\sum_i{\ln{p(w_i)}})}$$
The lower the perplexity, the more effective the model is.
We follow the standard procedure and sum over all the words.

\subsection{Comparative Models and Experimental Settings}


In order to demonstrate the effectiveness of TSLM, we compare our model with several baseline models, including Kneser-Ney 5-gram (KN-5)~\cite{mikolov2012context}, RNN based language model~\cite{Mikolov2010Recurrent}, Long Short-Term Memory network (LSTM) based language model~\cite{zaremba2014recurrent}, and RNN added Matrix of Softmax (MoS) language model~\cite{yang2018Breaking}. Models tagged with $\dagger$ indicate that they are reimplemented by ourselves.

\textbf{Kneser-Ney 5-gram (KN-5)}: It uses Kneser-Ney Smoothing on the $n$-gram language model (n=5)~\cite{chen1996empirical,mikolov2012context}. It is also the most representative statistical language model, and we consider it as a low-order tensor language model.

\textbf{RNN}: Recurrent neural network based language models~\cite{Mikolov2010Recurrent}, use a recurrent hidden layer to represent longer and variable length histories, instead of using fixed number of words to represent the context.

\textbf{LSTM}: LSTM neural network is an another variant of RNN structure. It allows to discover both long and short patterns in data and eliminates the problem of vanishing gradient by training RNN. LSTM approved themselves in various applications and it seems to be very promising course also for the field of language modeling~\cite{Soutner2013Application}.

\textbf{RNN+MoS}: A high-rank RNN language model~\cite{yang2018Breaking} breaking the softmax bottleneck, formulates the next-token probability distribution as a Matrix of Softmax (MoS), and improves the perplexities on the PTB and WikiText-2. It is the state-of-the-art softmax technique for solving probability distributions.

Since our focus is to show the effectiveness of language modeling in tensor space, we choose to set a relatively small scale network structure. Specifically, we set the same scale parameters for comparison experiments, i.e. 256/512 hidden size, 1 hidden layer, 20 batch size and 30/40 sequence length. Among them, the hidden size is equivalent to the tensor decomposition rank $r$
, and sequence length means the tensor order $n$ in our model.

\subsection{Experimental Results and Analysis}
Table~\ref{table:ptbwiki2} shows the results on PTB and WT2, respectively. 
From Table~\ref{table:ptbwiki2}, we could observe that our proposed TSLM achieves the lower perplexity, which reflects that TSLM outperform others. It is obvious that KN-5 method gets the worst performance. 
We choose KN-5 as a baseline, since it is a typical $n$-gram model and we prove that TSLM is a generalization of $n$-gram. 
Although the most advanced Kneser-Ney Smoothing is used, 
it still has not achieved good performance. The reason could be that statistic language model is based on the word frequency statistics and does not have the semantic advantages that word vectors in continuous space can satisfy.


The LSTM based language modeling can theoretically model arbitrarily long dependencies. The experimental results show that our model has achieved relatively better results than LSTM reimplemented by ourselves. Based on the MoS, RNN based language models achieve state-of-the-art results. To prove the effectiveness of our model using MoS, we compared our model to RNN+MoS model on the same parameters. The empirical results of our model have also been improved, which also illustrates the effectiveness of our model structure. 
\begin{figure}[t]
\centering
\includegraphics[width=7.8cm,height=4cm]{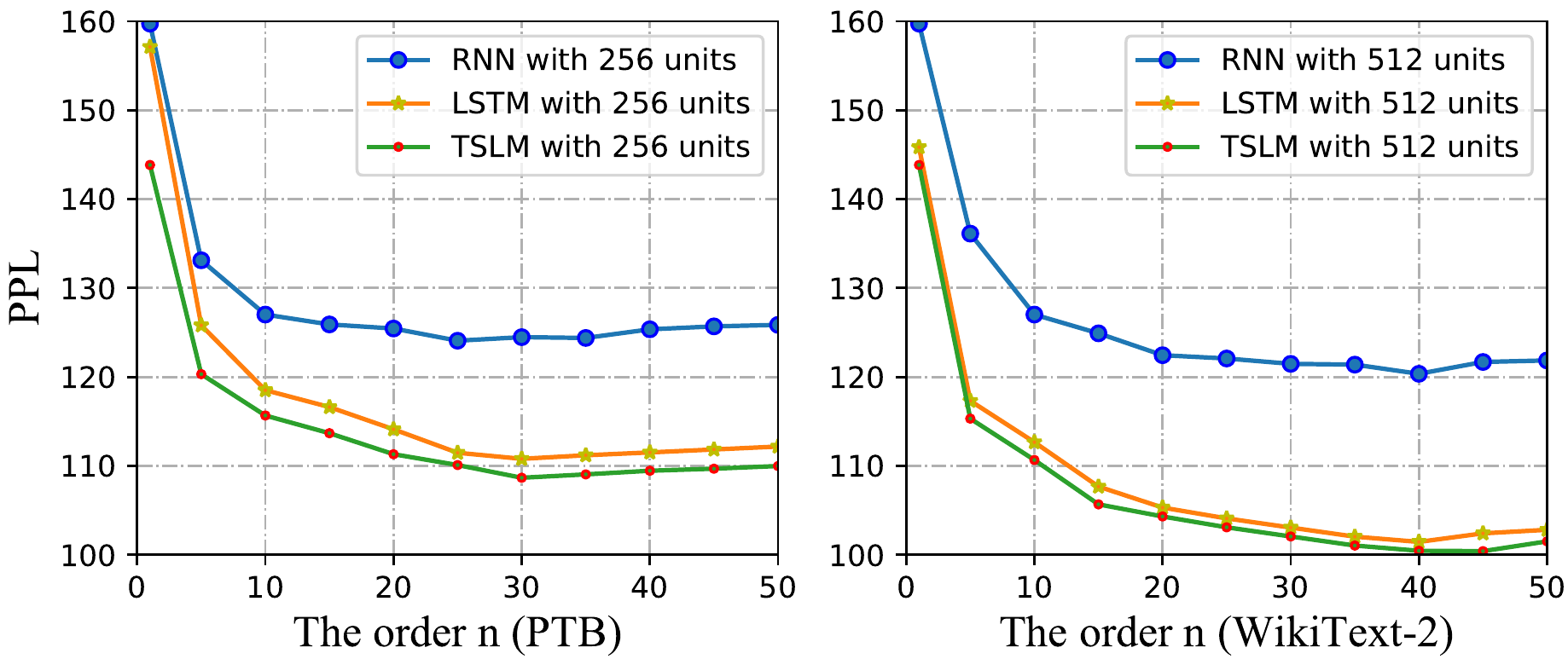}
\caption{Perplexity (PPL) with different max length of sentences in corpus.}
\label{fig:figure4}
\end{figure}

We have shown that TSLM is a generalized language model. In practice, its performance is better than RNN based language model. The reason could be that the non-linear operation $g$ in Eq.~\ref{eq:RNN} is an element-wise multiplication, which is capable of expressing more complex semantic information than a standard RNN structure with activation function using an addition operation.

Note that, the parameters $n$, $m$ and $r$ are crucial factors for modeling language in TSLM. The order $n$ of the tensor reflects the maximum sentence length. With the increase of the maximum sentence length, the performance of the model gradually increases. However, after a certain degree of growth, the expressive ability of the TSLM will reach an upper bound. The reason can be summarized as: The size of the corpus determines the size of the tensor space we need to model. The larger the order is, the larger the semantic space that TSLM can model and the more semantic information it can contain. As shown in Fig.~\ref{fig:figure4}, we can see that the model is optimal when $n$ equals 30 and 40 on PTB and WikiText datasets, respectively. There are similar experimental phenomena in RNN and LSTM. 

\begin{figure}[t]
\centering
\includegraphics[width=7.7cm,height=4cm]{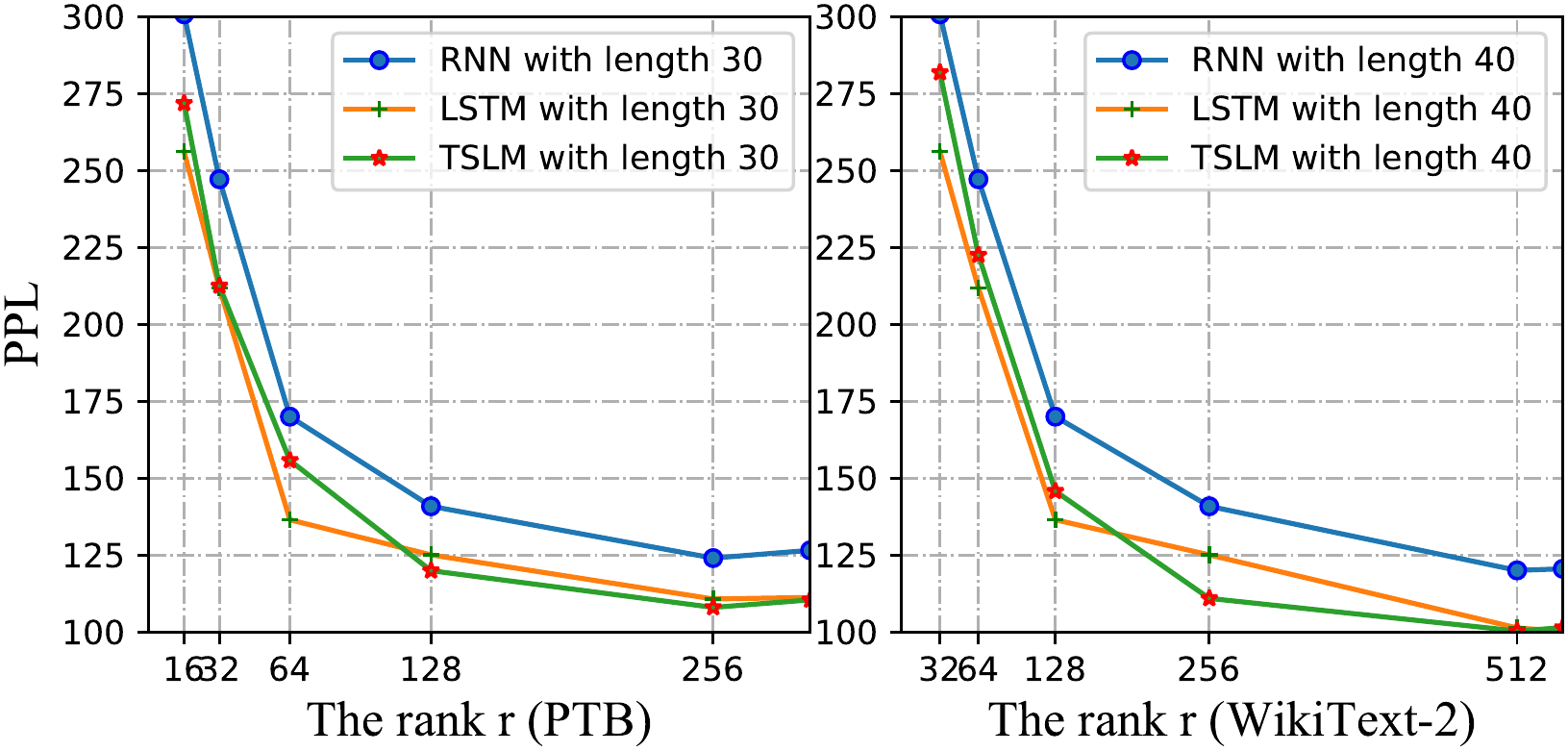}
\caption{Perplexity (PPL) with different hidden sizes.}
\label{fig:figure5}
\end{figure}

Other key factors that affect the capability of TSLM are the dimension of the orthogonal basis $m$ and the rank of tensor decomposition $r$, where each orthogonal basis denote the basic semantic meaning. The tensor decomposition is enough to extract the main features of the tensor $\mathcal{T}$ when $r=m$. They correspond to the word embedding size and hidden size in RNN or LSTM, and are usually set as the same value. We try to set the value of them as $[16,32,64,128,256,512,...]$. As shown in Fig.~\ref{fig:figure5}, we can see that the model is optimal when the decomposition rank $r$ equals 256 and 512 on PTB and WikiText datasets, respectively. These phenomena is mainly due to the saturation of semantic information in tensor space, which means that the number of the basic semantic meaning is finite with respect to the specific corpus.

\section{Conclusions and Future work}
\label{sect:7}
In this paper, we have proposed a language model based on tensor space, named Tensor Space Language Model (TSLM). We use the tensor networks to represent the high-order tensor for modeling language, and calculate the probability of a specific sentence by the inner product of tensors. Moreover, we have proved that TSLM is a generalization of $n$-gram language model and we can derive the recursive language model process from TSLM. Experimental results demonstrate the effectiveness of our model, compared with the standard RNN-based language model and LSTM-based language model on two typical datasets to evaluate language modeling. In the future, we will further explore the potential of tensor network for  modeling language in both theoretical and empirical directions.

\section{Acknowledgement}
This work is supported in part by the state key development program of China (grant No. 2017YFE0111900), Natural Science Foundation of China (grant No. U1636203, 61772363), and the European Union’s Horizon 2020 research and innovation programme under the Marie Skłodowska-Curie grant agreement No. 721321.
\bibliography{aaai2019}
\bibliographystyle{aaai}

\section{Appendix}
\subsection{Claim 1}
In our TSLM, when we set the dimension of vector space $m=|\mathrm{V}|$ and each word $w$ as an \textit{one-hot} vector, the probability of sentence $s$ consist of words $d_1,\ldots,d_n$ in vocabulary is the entry $\mathcal{T}_{d_1\ldots d_n}$ of tensor $\mathcal{T}$.
\subsection{Proof of Claim 1}
\begin{proof}
In our TSLM, when we set the dimension of vector space $m=|\mathrm{V}|$ and each word $w$ as an one-hot vector, the specific sentence $s$ will be represented as an one-hot tensor. The mixed representation $\bm{c}$ can be regarded as the total sampling distribution. The tensor inner product $\langle \bm{s},\bm{c}\rangle$ represents statistics probability that a sentence $\bm{s}$ appears in a language. Specifically, the word $w_i$ is an one-hot vector $\bm{w}_i=(0,\ldots,1,\ldots,0)$, and the vectors of any two different words are orthogonal (word vector itself is basis vector):
\begin{equation}
\langle\bm{w}_i,\bm{w}_j\rangle=\langle\bm{e}_i,\bm{e}_j\rangle=\left\{
             \begin{array}{lr}
             1,~~~& i=j   \\
             0,~~~& i\neq j
             \end{array}
\right.
\end{equation}

Firstly, for the sentence $s=(w_1,\ldots,w_n)$ with length $n$ is represented as $\bm{s}=\bm{w}_{1}\otimes\cdots\otimes\bm{w}_{n}$. It is an one-hot tensor, and the tensors of any two different sentences are orthogonal (when being viewed as the flatten vectors):
\begin{equation}
\begin{aligned}
\bm{s}_i&=\bm{w}_{i,1}\otimes\cdots\otimes\bm{w}_{i,n}\\
\bm{s}_j&=\bm{w}_{j,1}\otimes\cdots\otimes\bm{w}_{j,n}\\
\Rightarrow\langle\bm{s}_i,\bm{s}_j\rangle&=\prod_{k=1}^n\langle\bm{w}_{i,k},\bm{w}_{j,k}\rangle\\
&=\left\{
             \begin{array}{lr}
             1,~~~ \bm{w}_{i,k}=\bm{w}_{j,k},~\forall~{k}\in[n]\\
             0,~~~ \mathrm{otherwise}~&
             \end{array}
\right.\\
\Rightarrow\langle\bm{s}_i, \bm{c}\rangle&=\langle\bm{s}_i,\sum_jp_j\bm{s}_j\rangle=p_i
\end{aligned}
\end{equation}
Secondly, for the representations in TSLM, the orthogonal basis can be composed by $\{\bm{w}_d\}_{d=1}^{|\mathrm{V}|}$, and the sentence $s$ will be represented as:
\begin{equation}
\bm{s}=\sum_{d_1,\ldots,d_n=1}^{|\mathrm{V}|}\mathcal{A}_{d_1\ldots d_n}\bm{w}_{d_1}\otimes\cdots\otimes\bm{w}_{d_n}
\end{equation}
where
\begin{equation}
\mathcal{A}_{d_1\ldots d_n} = \left\{
             \begin{array}{lr}
             1,~~~ d_k=\mathrm{index}(w_k,\mathrm{V}),\forall~{k}\in[n]&  \\
             0,~~~ \mathrm{otherwise}& \\
             \end{array}
\right.
\end{equation}
which means tensor $\mathcal{A}$ is an one-hot tensor, and $\mathrm{index}$($w$,$\mathrm{V}$) means the index of $w$ in vocabulary $\mathrm{V}$. We have defined $\bm{c}:= \sum p_i\bm{s}_i$ as :
\begin{equation}
\bm{c}=\sum_{d_1,\ldots,d_n=1}^{|\mathrm{V}|}\mathcal{T}_{d_1\ldots d_n}\bm{w}_{d_1}\otimes\cdots\otimes\bm{w}_{d_n}
\end{equation}
Then, the probability of sentence $s_i$ is:
\begin{equation}
\begin{aligned}
p_i&=\langle\bm{s}_i, \bm{c}\rangle\\
&=\sum_{d_1,\ldots,d_n=1}^{|\mathrm{V}|}\mathcal{T}_{d_1\ldots d_n}\mathcal{A}_{d_1\ldots d_n}\\
&=\mathcal{T}_{d_1\ldots d_n},~~d_k=\mathrm{index}(w_k,\mathrm{V}),\forall~{k}\in[n]&\\
\end{aligned}
\end{equation}
Therefore, the probability of sentence $s$ consist of words $d_1,\ldots,d_n$ is $p(w_1\ldots w_n=d_1\ldots d_n) = \mathcal{T}_{d_1\ldots d_n}$. 
\end{proof}
\subsection{Claim 2}
In our TSLM, we define the word sequence $w_1^i=(w_1,w_2,\ldots,w_i)$ with length $i$ as:
\begin{equation}
\bm{w}_{1}^{i} := \bm{w}_{1}\otimes\cdots\otimes\bm{w}_{i}\otimes\bm{1}_{i+1}\otimes\cdots\otimes\bm{1}_n
\end{equation}
Then, the probability $p(w_1^i)$ can be computed as $p(w_1^i)=\langle \bm{w}_1^i,\bm{c}\rangle$.

\subsection{Proof of Claim 2}
\begin{proof}
According to the marginal distribution in probability theory and statistics, for two discrete random variables, the marginal probability function can be written as $p(X=x)$:
\begin{equation}
p(X=x)=\sum_yp(X=x,Y=y) 
\end{equation}
where $p(X=x,Y=y)$ is the joint distribution of two variables $X$ and $Y$. 

Firstly, in our TSLM, we can define the marginal distribution using the word variable $w$ as:
\begin{equation}
\begin{aligned}
p(w_i)&=\sum_{w_j\in\mathrm{V}}p(w_i,w_j)\\
p(w_1,\ldots,w_{n-1})&=\sum_{w_n\in\mathrm{V}}p(w_1,\ldots,w_{n-1},w_n)\\
\end{aligned}
\end{equation}
Secondly, the probability of the word sequence $w_1^i$ can be written as:
\begin{equation}
\label{p_w_1_i}
\begin{aligned}
&p(w_1^i)\\
=&p(w_1,\ldots,w_{i})\\
=&\sum_{w_{i+1},\cdots,w_n\in\mathrm{V}}p(w_1,\ldots,w_i,w_{i+1},\ldots,w_n)\\
=&\sum_{d_{i+1},\cdots,d_n=1}^{|\mathrm{V}|}\mathcal{T}_{d_1\ldots d_n},~~d_k=\mathrm{index}(w_k,\mathrm{V}),\forall~{k}\in[i]
\end{aligned}
\end{equation}
Then, the inner product $\langle\bm{w}_1^i,\bm{c}\rangle$ can be written as:
\begin{equation}
\label{innerproduct_w1i_c}
\begin{aligned}
&\langle\bm{w}_1^i,\bm{c}\rangle\\
=&\langle\bm{w}_1\otimes\cdots\otimes\bm{1},\sum_{d_{1},\cdots,d_n=1}^{|\mathrm{V}|}\mathcal{T}_{d_1\ldots d_n}\bm{w}_{d_1}\otimes\cdots\otimes\bm{w}_{d_n}\rangle\\
=&\sum_{d_{1},\cdots,d_n=1}^{|\mathrm{V}|}\mathcal{T}_{d_1\ldots d_n}\langle\bm{w}_1\otimes\cdots\otimes\bm{1},\bm{w}_{d_1}\otimes\cdots\otimes\bm{w}_{d_n}\rangle\\
=&\sum_{d_{1},\cdots,d_n=1}^{|\mathrm{V}|}\mathcal{T}_{d_1\ldots d_n}\prod_{j=1}^{i}\langle\bm{w}_j,\bm{w}_{d_j}\rangle\prod_{j=i+1}^{n}\langle\bm{1},\bm{w}_{d_j}\rangle\\
=&\sum_{d_{i+1},\cdots,d_n=1}^{|\mathrm{V}|}\mathcal{T}_{d_1\ldots d_n},~~d_k=\mathrm{index}(w_k,\mathrm{V}),\forall~{k}\in[i]
\end{aligned}
\end{equation}
Therefore, we derive that the probability $p(w_1^i)$ can be computed as $p(w_1^i)=\langle \bm{w}_1^i,\bm{c}\rangle$ according to Eq.~\ref{p_w_1_i} and \ref{innerproduct_w1i_c}.
\end{proof} 
\end{document}